\documentclass{article}

\usepackage[final]{corl_2017} 

\usepackage{multirow}
\usepackage{times}
\usepackage{array}   
\newcolumntype{C}{>{$}c<{$}} 
\usepackage{float}
\usepackage[numbers]{natbib}
\usepackage{multicol}
\usepackage{caption}
\usepackage{graphicx}
\usepackage{booktabs}
\usepackage{color}
\usepackage{amsmath}
\usepackage[english, status=draft]{fixme}
\fxusetheme{color}
\usepackage[font=small,labelfont=bf]{caption}

\DeclareMathOperator*{\argmin}{arg\,min}

\title{Toward Geometric Deep SLAM}

\author{
  Daniel DeTone\\
  Magic Leap, Inc.\\
  Sunnyvale, CA\\
  \texttt{ddetone@magicleap.com} \\
  \And
  Tomasz Malisiewicz \\
  Magic Leap, Inc. \\
  Sunnyvale, CA\\
  \texttt{tmalisiewicz@magicleap.com} \\
  \And
  Andrew Rabinovich \\
  Magic Leap, Inc. \\
  Sunnyvale, CA\\
  \texttt{arabinovich@magicleap.com} \\
}

\begin{document}
\maketitle

\begin{abstract}
We present a point tracking system powered by two deep convolutional neural networks. The first network, MagicPoint, operates on single images and extracts salient 2D points. The extracted points are ``SLAM-ready'' because they are by design isolated and well-distributed throughout the image. We compare this network against classical point detectors and discover a significant performance gap in the presence of image noise. As transformation estimation is more simple when the detected points are geometrically stable, we designed a second network, MagicWarp, which operates on pairs of point images (outputs of MagicPoint), and estimates the homography that relates the inputs. This transformation engine differs from traditional approaches because it does not use local point descriptors, only point locations. Both networks are trained with simple synthetic data, alleviating the requirement of expensive external camera ground truthing and advanced graphics rendering pipelines. The system is fast and lean, easily running 30+ FPS on a single CPU.
\end{abstract}



\section{Introduction}
\vspace{-.1in}
Much of deep learning success in computer vision tasks such as image categorization and object detection stems from the availability of large annotated databases like ImageNet and MS-COCO. However, for SLAM-like pose tracking and reconstruction problems, there instead exists a fragmented ecosystem of smaller device-specific datasets such as the Freiburg-TUM RGBD Dataset \cite{sturm12} based on the Microsoft Kinect, the EuRoC drone/MAV dataset \cite{burri16} based on stereo vision cameras and IMU, and the KITTI driving dataset~\cite{geiger12}.

We frequently ask ourselves: \emph{what would it take to build an ImageNet for SLAM?} Obtaining accurate ground-truth pose measurements for a large number of environments and scenarios is difficult. Getting accurate alignment between ground-truthing sensors and a standard set of Visual SLAM sensors takes significant effort; it is expensive and is difficult to scale across variations in cameras. Category labeling in a crowd-sourced or pay-per-label Amazon Mechanical Turk fashion, as is commonly done for ImageNet-like datasets, suddenly seems a lot more fun.

Photorealistic rendering is potentially useful, as all relevant geometric variables for SLAM tasks can be recorded with 100\% accuracy. Benchmarking SLAM on photorealistic sequences makes a lot of sense, but training on such rendered images often suffers from domain adaptation issues. Our favorite deep nets seem to overfit. Datasets have been created with this intent in mind, but as the research community always demands results on real-world datasets, the benefit of photorealistic rendering for automatic SLAM training is still a dream. Since a public ImageNet-scale SLAM dataset does not exist today and photorealistic rendering brings its own set of new problems, how are we to embrace the data-driven philosophy of deep learning while building an end-to-end Deep SLAM system? Our proposed solution comes from a couple of key insights.

First, recent ego-motion estimation has shown that it is possible to train deep convolutional neural networks on the task of image prediction. Compared to direct supervision (i.e., regressing to the ground truth 6 DoF pose), supervision for frame prediction comes ``for free.'' This new insight that the move away from strong-supervision might bear more fruits is rather welcoming for SLAM, an ecosystem already plagued with a fragmentation of datasets. Systems such as \cite{zhou17} perform full frame prediction, and we will later see that our flavor of the prediction problem is more geometric, as we focus on \emph{geometric consistency}.

\begin{figure*}
\centering
\includegraphics[width=\textwidth]{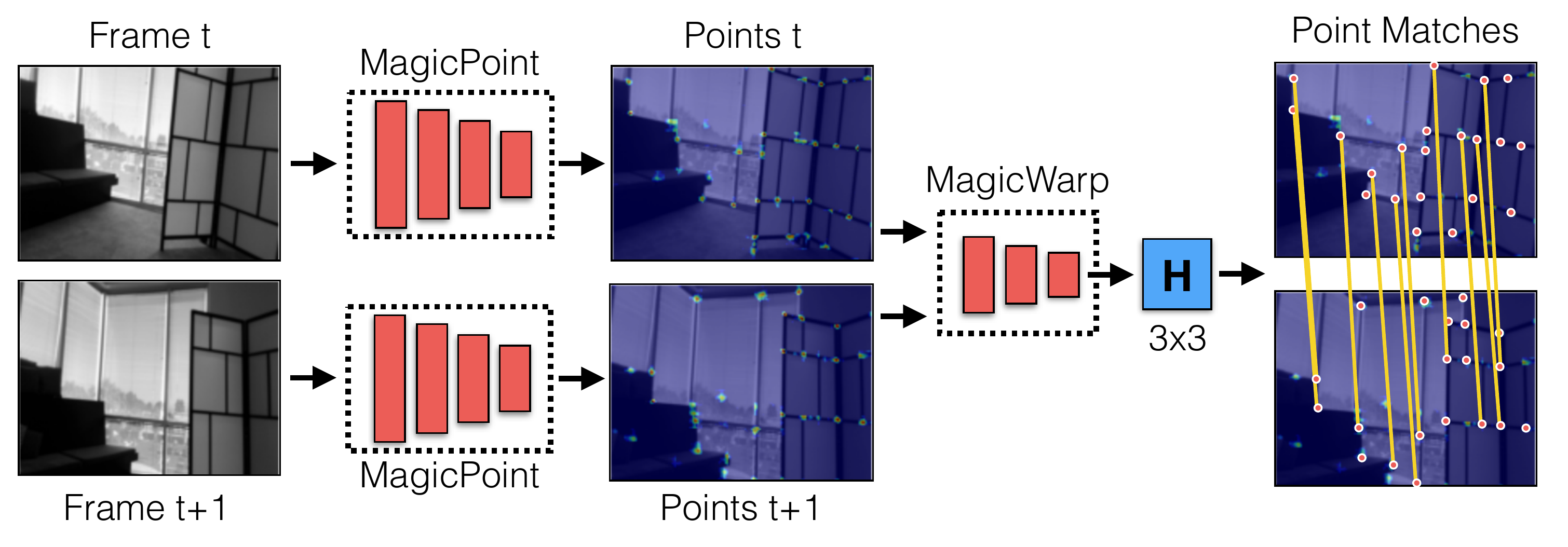}
\caption{{\bf Deep Point-Based Tracking Overview.} Pairs of images are processed by a convolutional neural network called MagicPoint which is trained to detect salient corners in the image. The resulting point images are then processed together by MagicWarp (another convolutional neural network) to compute a homography $H$ which relates the points in the input images.
\label{fig:overview}}
\vspace{-.1in}
\end{figure*}

Second, SLAM models must be lean or they will not run at a large scale on embedded platforms such as those in robotics and augmented reality. Our desire to focus on geometric consistency as opposed to full frame prediction comes from a dire need to deploy such systems in production. While it is fulfilling to watch full frame predictions made by a deep learning system, we already known from previous successes in SLAM (e.g., \cite{klein2007} and \cite{mur2015}) that predicting/aligning points is sufficient for metric-level pose recovery. So why solve a more complex full frame prediction task than is necessary for SLAM?


\vspace{-.1in}
\section{Related Work}
\vspace{-.1in}
\label{sec:related}

Individual components of SLAM systems have recently been tackled with supervised deep learning methods. The feature point detection and description stage was tackled in the work of \cite{yi16}, where a convolutional neural network was trained using image patches filtered through a classical Structure From Motion pipeline. Transformation estimation was shown to be done successfully by CNNs in \cite{detone16}, where a deep network was trained on a large dataset of warped natural images. The transformation estimation done in this work was direct, meaning that the convolutional neural network directly mapped pairs of images to their transforms. The work of \cite{fischer2015} tackled dense optical flow. The problem of camera localization was also tackled with a CNN in \cite{kendall15}, where a network was trained to learn a mapping from images to absolute 6DOF poses.  A deep version of the RANSAC algorithm was presented in \cite{brachmann16}, where a deep network was trained to learn a robust estimator for camera localization. There have also been many works such as \cite{gomez15} in the direction of deep relocalization via metric learning, where two images with similar pose are mapped to a similar point on an embedding produced by a deep network.

A series of works have also tackled multiple SLAM problems concurrently. In \cite{ummenhofer16}, a CNN was trained to simultaneous estimate the depth and motion of a monocular camera pair. Interestingly, the models did not work well when the network was trained on the two tasks separately, but worked much better when trained jointly. This observation is consistent with natural regularization of multi task learning.  This approach relies on supervised training data for both motion and depth cues.

There is also a new direction of research at the intersection of deep learning and SLAM where no or very few ground truth measurements are required. By formulating loss functions to maximize photometric consistency, the works of \cite{zhou17} and \cite{godard17} showed an ego-motion and depth estimation system based on image prediction. Work in this direction show that there is much promise in moving away from strong supervision.

\section{Deep Point-Based Tracking Overview}
\vspace{-.1in}
\label{sec:overview}

The general architecture of our Deep Point-Based Tracking system is shown in Figure~\ref{fig:overview}. There are two convolutional neural networks that perform the majority of computation in the tracking system: MagicPoint and MagicWarp. We discuss these two models in detail below.

\begin{figure*}
\centering
\includegraphics[width=\textwidth]{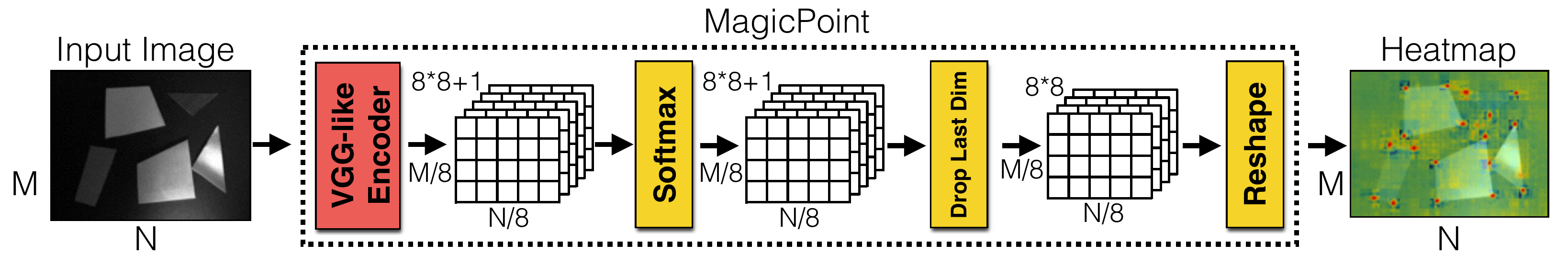}
\vspace{-.1in}
\caption{{\bf MagicPoint architecture.} The MagicPoint network operates on grayscale images and outputs a ``point-ness'' probability for each pixel. We use a VGG-style encoder combined with an explicit decoder. Each spatial location in the final $15$x$20$x$65$ tensor represents a probability distribution over a local 8x8 region plus a single dustbin channel which represents no point being detected $(8*8+1=65)$. The network is trained using a standard cross entropy loss, using point supervision from the 2D shape renderer (see examples in Figure~\ref{fig:mpdata}). \label{fig:mpnet}}
\vspace{-.1in}
\end{figure*}

\subsection{MagicPoint Overview}

\textbf{MagicPoint Motivation.} The first step in most sparse SLAM pipelines is to detect stable 2D interest point locations in the image. This step is traditionally performed by computing corner-like gradient response maps such as the second moment matrix \cite{harris1988} or difference of Gaussians \cite{lowe2004} and detecting local maxima. The process is typically repeated at various image scales. Additional steps may be performed to evenly distribute detections throughout the image, such as requiring a minimum number of corners within an image cell \cite{mur2015}. This process typically involves a high amount of domain expertise and hand engineering, which limits generalization and robustness. Ideally, interest points should be detected in high sensor noise scenarios and low light. Lastly, we should get a confidence score for each point we detect that can be used to help reject spurious points and up-weigh confident points later in the SLAM pipeline.

\textbf{MagicPoint Architecture.} We designed a custom convolutional network architecture and training data pipeline to help meet the above criteria. Ultimately, we want to map an image $I$ to a point response image $P$ with equivalent resolution, where each pixel of the output corresponds to a probability of ``corner-ness'' for that pixel in the input. The standard network design for dense prediction involves an encoder-decoder pair, where the spatial resolution is decreased via pooling or strided convolution, and then upsampled back to full resolution via upconvolution operations, such as done in \cite{vijay15}. Unfortunately, upsampling layers tend to add a high amount of compute, thus we designed the MagicPoint with an explicit decoder\footnote{Our decoder has no parameters, and is known as ``sub-pixel convolution''~\cite{Shi2016} or ``depth to space'' inside TensorFlow.} to reduce the computation of the model. The convolutional neural network uses a VGG style encoder to reduce the dimensionality of the image from 120x160 to 15x20 cell grid, with 65 channels for each spatial position. In our experiments we chose the QQVGA resolution of 120x160 to keep the computation small. The 65 channels correspond to local, non-overlapping 8x8 grid regions of pixels plus an extra dustbin channel which corresponds to no point being detected in that 8x8 region. The network is fully convolutional, using 3x3 convolutions followed by BatchNorm normalization and ReLU non-linearity. The final conv layer is a 1x1 convolution and more details are shown in Figure~\ref{fig:mpnet}.

\textbf{MagicPoint Training.} What parts of an image are interest points? They are typically defined by computer vision and SLAM researchers as uniquely identifiable locations in the image that are stable across a variety of viewpoint, illumination, and image noise variations. Ultimately, when used as a preprocessing step for a Sparse SLAM system, they must detect points that work well for a given SLAM system. Designing and choosing hyper parameters of point detection algorithms requires expert and domain specific knowledge, which is why we have not yet seen a single dominant point extraction algorithm persisting across many SLAM systems.

\begin{figure}[h]
\centering
\includegraphics[width=\textwidth]{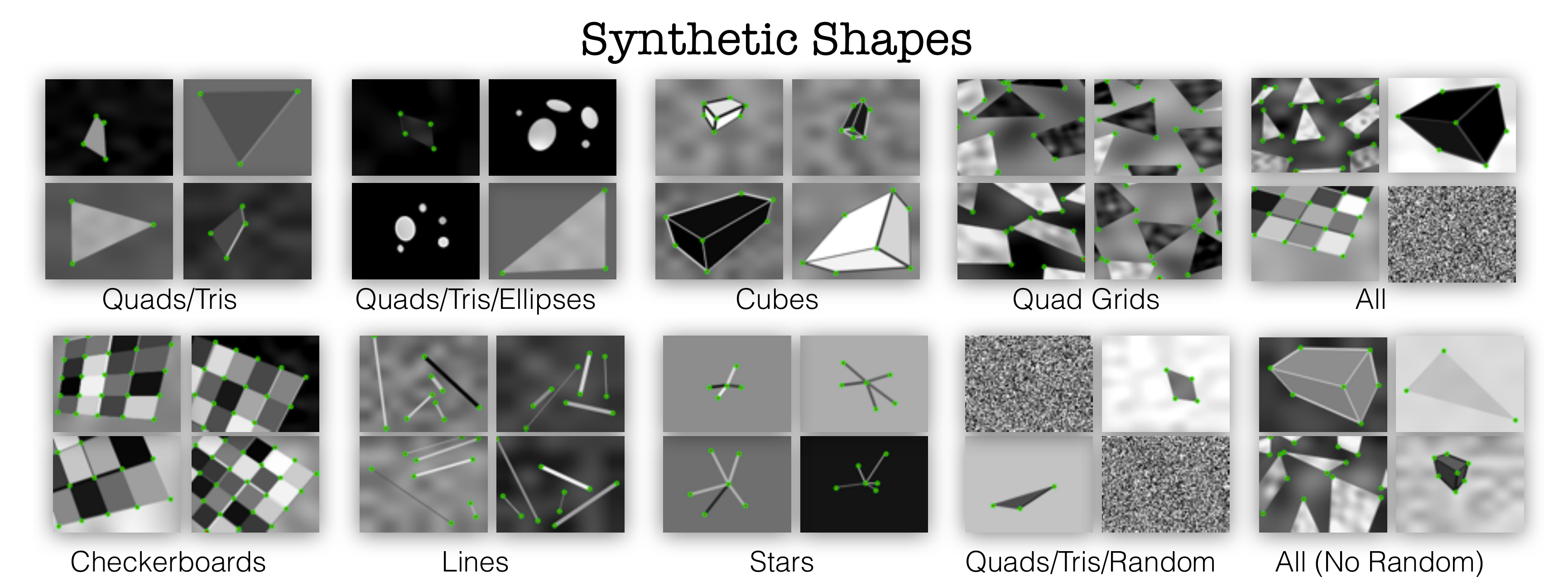}
\caption{{\bf Synthetic Shapes Dataset.} The {\tt Synthetic Shapes} dataset consists of rendered triangles, quadrilaterals, lines, cubes, checkerboards, and stars each with ground truth corner locations. It also includes some negative images with no ground truth corners, such as ellipses and random noise images. \label{fig:mpdata}}
\end{figure}

There is no large database of interest point labeled images that exists today. To avoid an expensive data collection effort, we designed a simple renderer based on available OpenCV~\cite{opencv} functions. We render simple geometric shapes such as triangles, quadrilaterals, stars, lines, checkerboards, 3D cubes, ellipses and random noise. For each image we know the ground truth corner locations. See Figure~\ref{fig:mpdata} for examples from our synthetic shapes renderer. Note the 2D ground truth locations need not correspond to local, high-gradient intersections of edges in the image, but can instead correspond to other low-level cues which require a larger local receptive field. We also trained MagicPoint networks that detect non-corner interest points such ellipse centers, 2D polygon face centers, and midpoints along edges. For simplicity, we only train on corners in this paper.

Once the shapes are rendered, we apply homographic warping to each image to augment the number of training examples and we apply high amounts noise in the form of brightness changes, shadows, blurring, Gaussian noise, and speckle noise. See Figure~\ref{fig:mpeval_noise_type} for examples of the noise applied during training. The data is generated on the fly and no example is seen by the network twice. The network is trained using a standard cross entropy loss after the logits for each cell in the 15x20 grid are passed through a softmax function.

\subsection{MagicWarp Overview}

\textbf{MagicWarp Motivation.} Our second network, MagicWarp, produces a homography given a pair of point images as produced by Magic Point. Once the homography is computed, the points in one image are transformed into the other and the point correspondence is computed by assigning correspondence to close neighbors. In doing so, MagicWarp estimates correspondence in image pairs without interest point descriptors. Once correct correspondences are established, it is straightforward to compute 6DOF relative pose $R$s and $t$s using either a homography matrix decomposition for planar scenes or a fundamental matrix decomposition for non-planar scenes, assuming the camera calibration matrix $K$ is known. By designing the network to operate on (the space of point images $\times$ the space of relative poses) instead of (the space of all images $\times$ the space of relative poses), we do not have to worry about illumination, shadows, and textures. We no longer rely on photometric consistency assumption to hold. Plus, by reducing the problem dimensionality, the transformation estimation model can be small and efficient.

\begin{figure*}
\centering
\includegraphics[width=\textwidth]{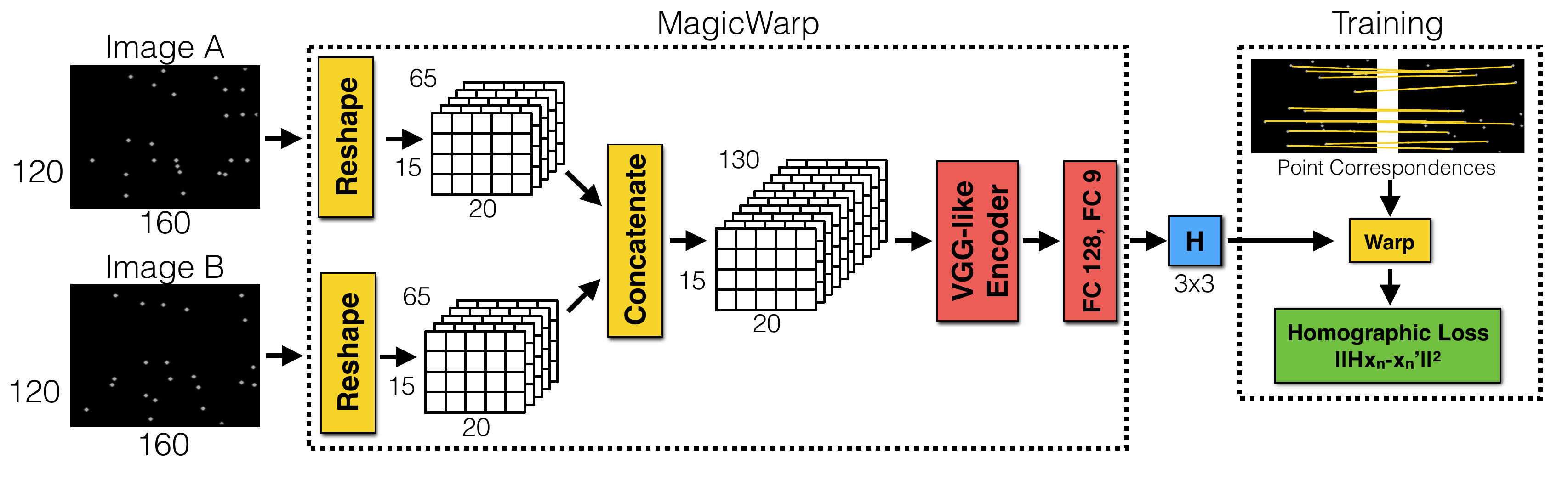}
\caption{{\bf MagicWarp architecture.} Pairs of binary point images are concatenated and then fed through a standard VGG-style encoder. The 3x3 homography H is output by a fully connected layer. H is then normalized such that its bottom right element is one. The loss is computed by warping points with known correspondence from one image into the other and measuring their distance to the ground truth correspondences. \label{fig:pbhnet}}
\end{figure*}

\textbf{MagicWarp Architecture.} MagicWarp is designed to operate directly on the point detections outputs from MagicWarp (although it can operate on any traditional point detector). We found that the model works well on pairs of the semi-dense 15x20x65 images. At this small spatial resolution the network uses very little compute. After channel-wise concatenation of the inputs to form an input of size 15x20x130, there is a VGG style encoder consisting of 3x3 convolutions, max-pooling, BatchNorm and ReLU activations, followed by two fully connected layers which output the 9 values of the 3x3 homography $H$. See Figure~\ref{fig:pbhnet} for more details. Note that MagicWarp can be applied iteratively, by using the network's first predicted $H_1$, applying it to one of the inputs, and computing a second $H_2$, yielding a final $H = H_1*H_2$, which improves results. For simplicity we do not apply MagicWarp iteratively in this paper.

\begin{figure*}
\centering
\includegraphics[width=\textwidth]{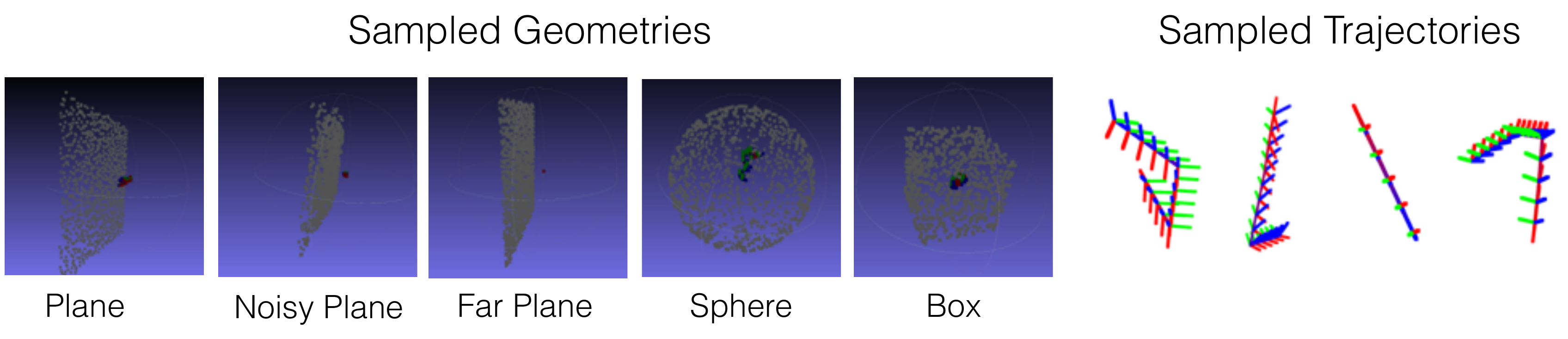}
\caption{{\bf MagicWarp data generation.} To generate 2D point set pairs, we create 3D point clouds of 3D geometries and render them to virtual cameras governed by simple 3D trajectories.\label{fig:phdata}}
\end{figure*}

\textbf{MagicWarp Training.} To train the MagicWarp network, we generate millions of examples of point clouds rendered into two virtual cameras. The point clouds are generated from simple 3d geometries, such as planes, spheres and cubes. The positions of the two virtual cameras are sampled from random trajectories which consist of piece-wise linear translation and rotations around random axes, as shown in Figure~\ref{fig:phdata}. We randomly sample camera pairs which have at least 30\% visual overlap. Once the points are projected into the two camera frames, we apply point input dropout, to improve the network's robustness to spurious and missing point detections. We found that randomly dropping 50\% of the matches and randomly dropping 25\% of the points independently works well.

The loss is computed by measuring the Euclidean distance between the correct matches. For $N$ matches in the left point image, each point $x_n$ is multiplied by the the predicted $H$ and compared to its match location in the right point image $x_{n}'$, shown in Equation~\ref{eqn:homography}.

\begin{equation}
L_{MagicWarp} = \sum_{n=1}^{N} {\lVert Hx_{n} - x_{n}' \rVert}^2
\label{eqn:homography}
\end{equation}

We found that care must be taken to train the network to directly output the 3x3 matrix. Training worked best when the final FC layer bias is initialized to output the identity matrix, when the coordinates of the homography H are normalized to the range $[-1,1]$, and when the $H$ quantity is normalized such that the bottom right element is one, since the homography $H$ has eight degrees of freedom and nine elements.

\section{MagicPoint Evaluation}
We evaluate the MagicPoint component of our system against traditional corner detection baselines like the FAST~\cite{rosten2006} corner detector, the Harris~\cite{harris1988} corner detector, and the ``Good Features to Track'' or Shi~\cite{shi1994} corner detector. For a thorough evaluation of classical corner detectors, see~\cite{gauglitz11}. The deep baselines are a small (MagicPointS, 81KB) and large version (MagicPointL, 3.1MB) of MagicPoint where the larger version was trained with corner-related sidetasks and a bigger network.

The detectors are evaluated on both synthetic and real image data. Both types of data consists of simple geometry that a human could easily label with the ground truth corner locations. While one could not build a fully functional SLAM system based on detectors which only work in these scenarios, we expect a good point detector to easily detect the correct corners in these scenarios. The added benefit of images with ground truth corner locations is that we can more rigorously analyze detector performance. In fact, we were surprised at how difficult the simple geometries were for the classical point detectors.

\subsection{Evaluation Measures}
\label{sec:measures}

\textbf{Corner Detection Average Precision.} We compute Precision-Recall curves and the corresponding Area-Under-Curve (also known as Average Precision), the pixel location error for correct detections, and the repeatability rate. For corner detection, we use a threshold $\varepsilon=4$ to determine if a returned point location $\mathbf{x}$ is correct relative to a set of $K$ ground-truth corners $\{\mathbf{ {\hat x}}_1, \dots, \mathbf{ {\hat x}}_K \}$. We define the correctness as follows:
\begin{equation}
  \verb+Corr+(\mathbf{x}) = (\min_{j} ||\mathbf{x} - {\bf {\hat x}}_j||) \leq \varepsilon
  \label{eqn:correct-detection}
\end{equation}
The precision recall curve is created by varying the detection
confidence and summarized with a single number, namely the Average Precision (which ranges from $0$ to $1$), and larger AP is better.

\textbf{Corner Localization Error.} To complement the AP analysis, we compute the corner localization error, but solely for the correct detections. We define the Localization Error as follows:
\begin{equation}
  \verb+LE+ = \frac{1}{N} \sum_{i : \verb+Corr+(\mathbf{x}_i)}
  \min_{j \in \{1,\dots,K\}}||\mathbf{x}_i - \mathbf{ {\hat x}}_j||
  \label{eqn:mse}
\end{equation}
The Localization Error is between $0$ and $\varepsilon$, and lower LE is better.

\textbf{Repeatability.} We compute the repeatability rate, which is the probability that a point gets detected in the next frame. We compute sequential repeatability (between frame $t$ and $t+1$ only). For repeatability, we also need a notion of correctness that relies on a pixel distance threshold. We use $\varepsilon=2$ for the threshold between points. Let's assume we have $N_1$ points in the first image and $N_2$ points in the second image. We define correctness for repeatability experiments as follows:

\begin{equation}
  \verb+Corr+(\mathbf{x}_i) = (\min_{j \in \{1,\dots,N_2\}} ||\mathbf{x}_i - {\bf {\hat x}}_j||) \leq \varepsilon
  \label{eqn:repeatability-detection}
\end{equation}
Repeatability simply measures the probability that a point is detected in the second image.
\begin{equation}
  \verb+Rep+ = \frac{1}{N_1+N_2} (\sum_{i} \verb+Corr+({\bf x}_i) + \sum_{j} \verb+Corr+({\bf x}_j))
\end{equation}
For each sequence of images, we want to create a single scalar Repeatability number. We first create a Repeatability vs Number of Detections curve, then find the point of maximum repeatability. When summarizing repeatability with a single number, we use the point of maximum repeatability, and report the result as repeatability@$N$ where $N$ is the average number of detections at the point of maximum repeatability.

\subsection{Results on Synthetic Shapes Dataset}

We created an evaluation dataset with our synthetic shapes generator to determine how well our detector is able to localize simple corners. There are 10 categories of images, shown in Figure~\ref{fig:mpdata}.

\begin{figure*}
\centering
\includegraphics[width=\textwidth]{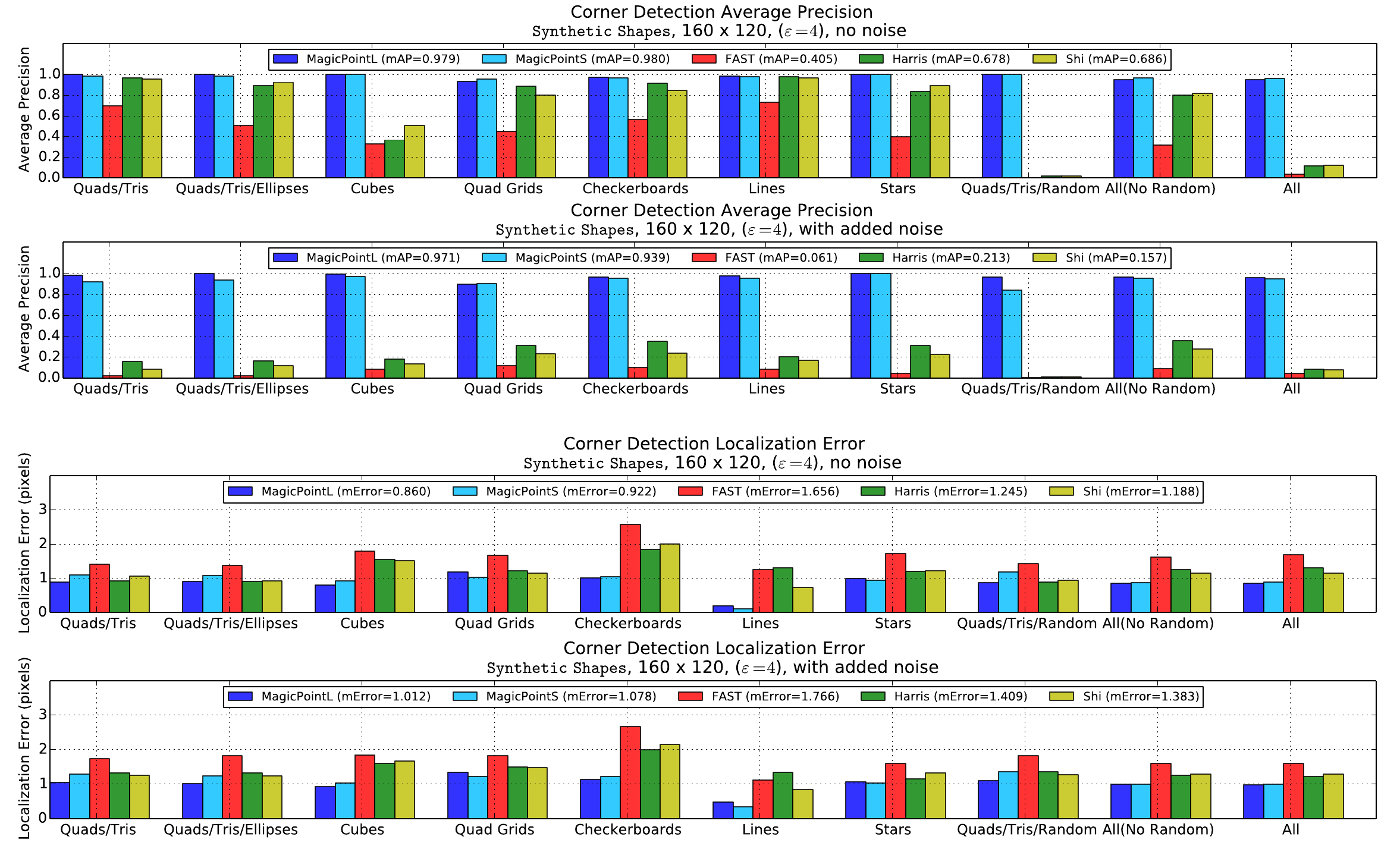}
\caption{{\bf Synthetic Shapes Results Plot.} These plots report Average Precision and Corner Localization Error for each of the 10 categories in the {\tt Synthetic Shapes} dataset with and without noise. The sequences with ``Random'' inputs are especially difficult for the classical detectors. \label{fig:mpeval}}
\vspace{-.1in}
\end{figure*}

\textbf{Mean Average Precision and Mean Localization Error.} For each category, there are 1000 images sampled from the synthetic shapes generator. We compute Average Precision and Localization Error with and without added imaging noise. A summary of the per category results are shown in Figure~\ref{fig:mpeval} and the mean results are shown in Table~\ref{table:mptable_ss}. The MagicPoint detectors outperform the classical detectors in all categories and in the mean. There is a significant performance gap in mAP in all categories in the presence of noise.

\begin{figure}
\centering
\includegraphics[width=\textwidth]{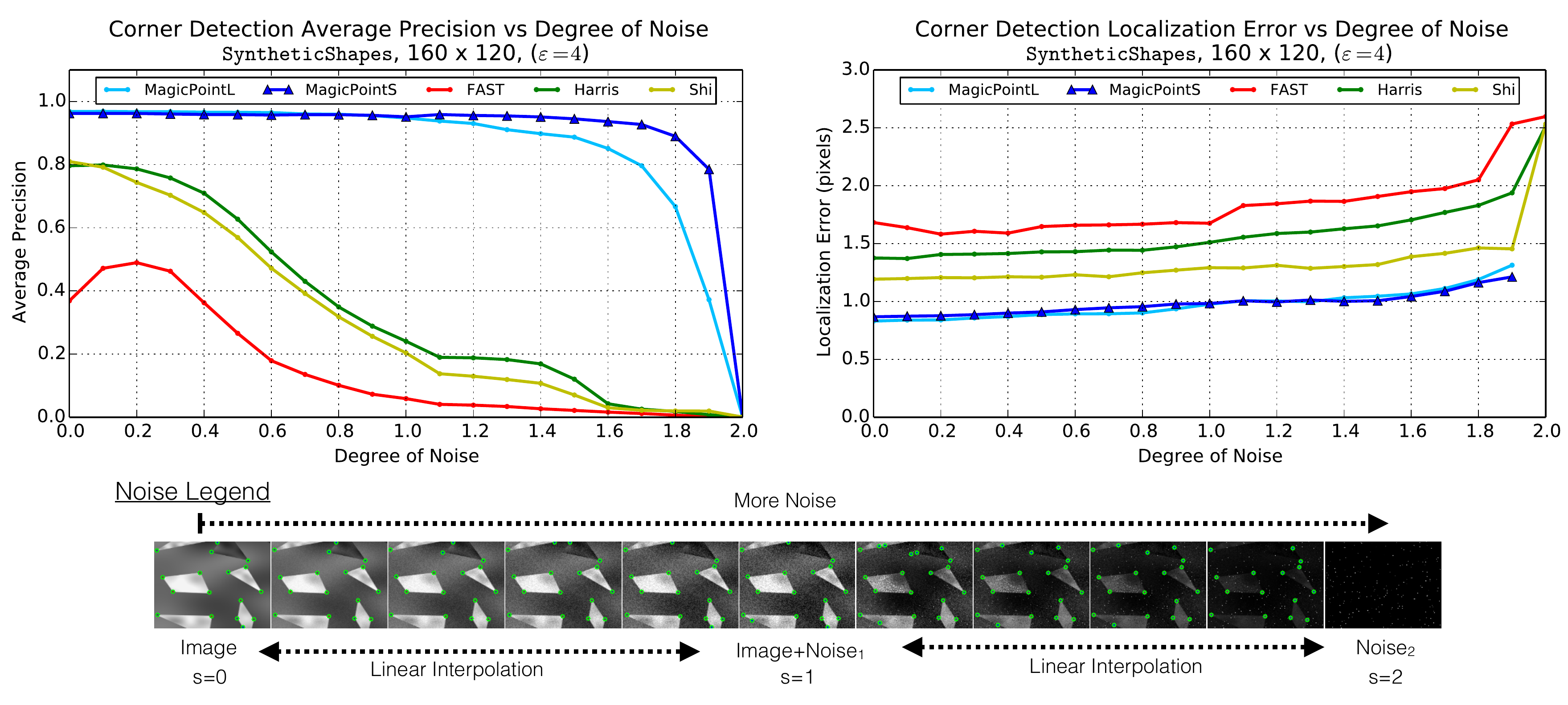}
\caption{{\bf Synthetic Shapes Effect of Noise Magnitude.} Two versions of MagicPoint are compared to three classical point detectors on the {\tt Synthetic Shapes} dataset (shown in Figure~\ref{fig:mpdata}). The MagicPoint models outperform the classical techniques in both metrics, especially in the presence of image noise. \label{fig:mpeval2}}
\end{figure}

\textbf{Effect of Noise Magnitude.} Next we study the effect of noise more carefully by varying its magnitude. We were curious if the noise we add to the images is too extreme and unreasonable for a point detector. To test this hypothesis, we linearly interpolate between the clean image ($s=0$) and the noisy image ($s=1$). To push the detectors to the extreme, we also interpolate between the noisy image and random noise ($s=2$). The random noise images contain no geometric shapes, and thus produce an mAP score of $0.0$ for all detectors. An example of the varying degree of noise and the plots are shown in Figure~\ref{fig:mpeval2}.

\begin{figure}
\centering
\includegraphics[width=\textwidth]{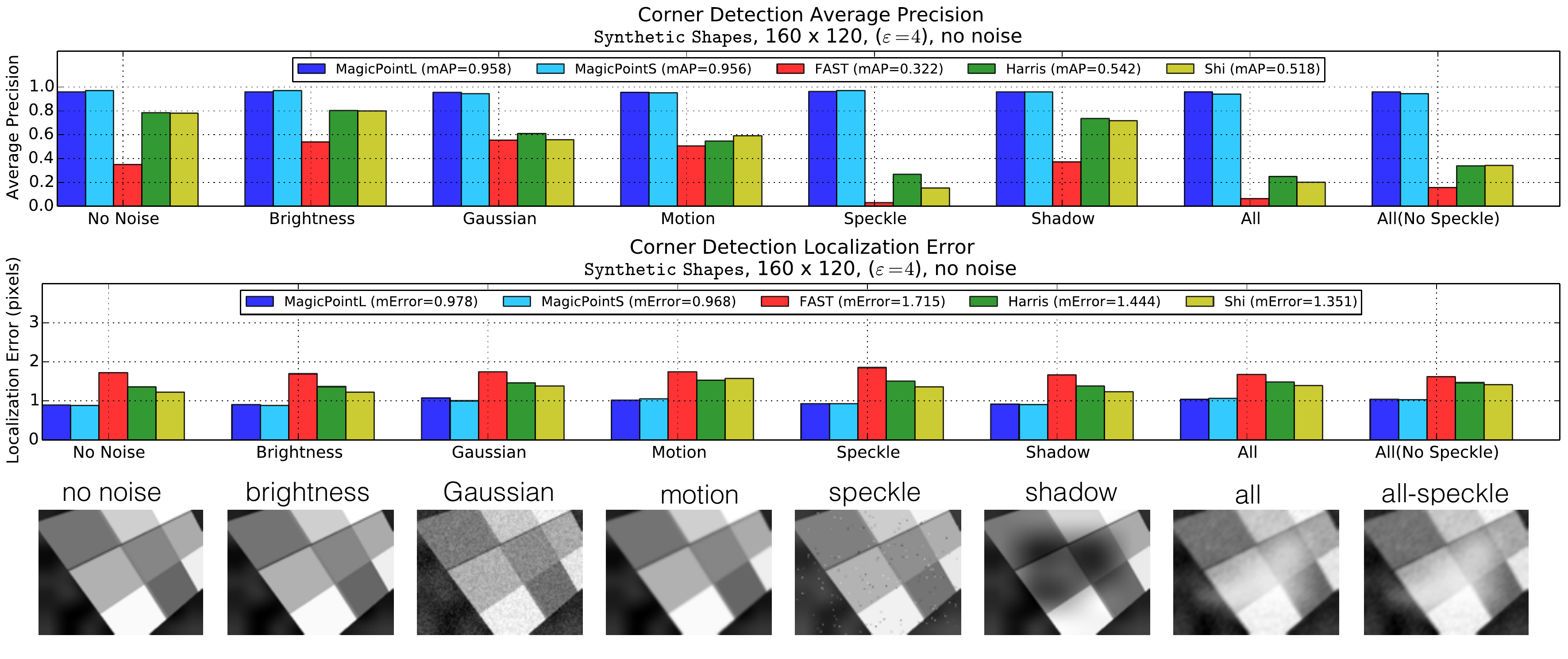}
\caption{{\bf Synthetic Shapes Effect of Noise Type.} The detector performance is broken down by noise category. Speckle noise is particularly difficult for traditional detectors. \label{fig:mpeval_noise_type}}
\end{figure}

\textbf{Effect of Noise Type.} We categorize the noise we apply into eight categories. We study the effect of each of these noise types individually to better understand which has the biggest effect on the point detectors. Speckle noise is particularly difficult for traditional detectors. Results are summarized in Figure~\ref{fig:mpeval_noise_type}.

\begin{table}
\begin{center}
 \begin{tabular}{| l | l | c c c c c|}
 \hline
 Metric & Noise & MagicPointL & MagicPointS & FAST & Harris & Shi\\
 \hline
 mAP & no noise & 0.979 & \textbf{0.980} & 0.405 & 0.678 & 0.686 \\
 \hline
 mAP & noise & \textbf{0.971} & 0.939 & 0.061 & 0.213 & 0.157 \\
 \hline
 MLE & no noise & \textbf{0.860} & 0.922 & 1.656 & 1.245 & 1.188 \\
 \hline
 MLE & noise & \textbf{1.012} & 1.078 & 1.766 & 1.409 & 1.383 \\
 \hline
\end{tabular}
\end{center}
\caption{\textbf{Synthetic Shapes Table Results.} Reports the mean Average Precision (mAP, higher is better) and Mean Localization Error (MLE, lower is better) across the 10 categories of images on the Synthetic Shapes dataset. Note that MagicPointL and MagicPointS are relatively unaffected by imaging noise.}
\label{table:mptable_ss}
\end{table}

\subsection{Results on 30 Static Corners Dataset}

We next evaluate MagicPoint on real data. We chose scenes with simple geometry so that the ground truth corner locations can be easily labeled by a human. These sequences are about 1-2 minutes in length and are recorded using a static, commodity webcam. Since the camera is static, we only label the first frame with ground truth corner locations and propagate the labels to all the other frames in the sequence. Throughout each sequence, we vary the lighting conditions using a hand-held point source light and overall room lighting.

\begin{figure}
\centering
\includegraphics[width=\textwidth]{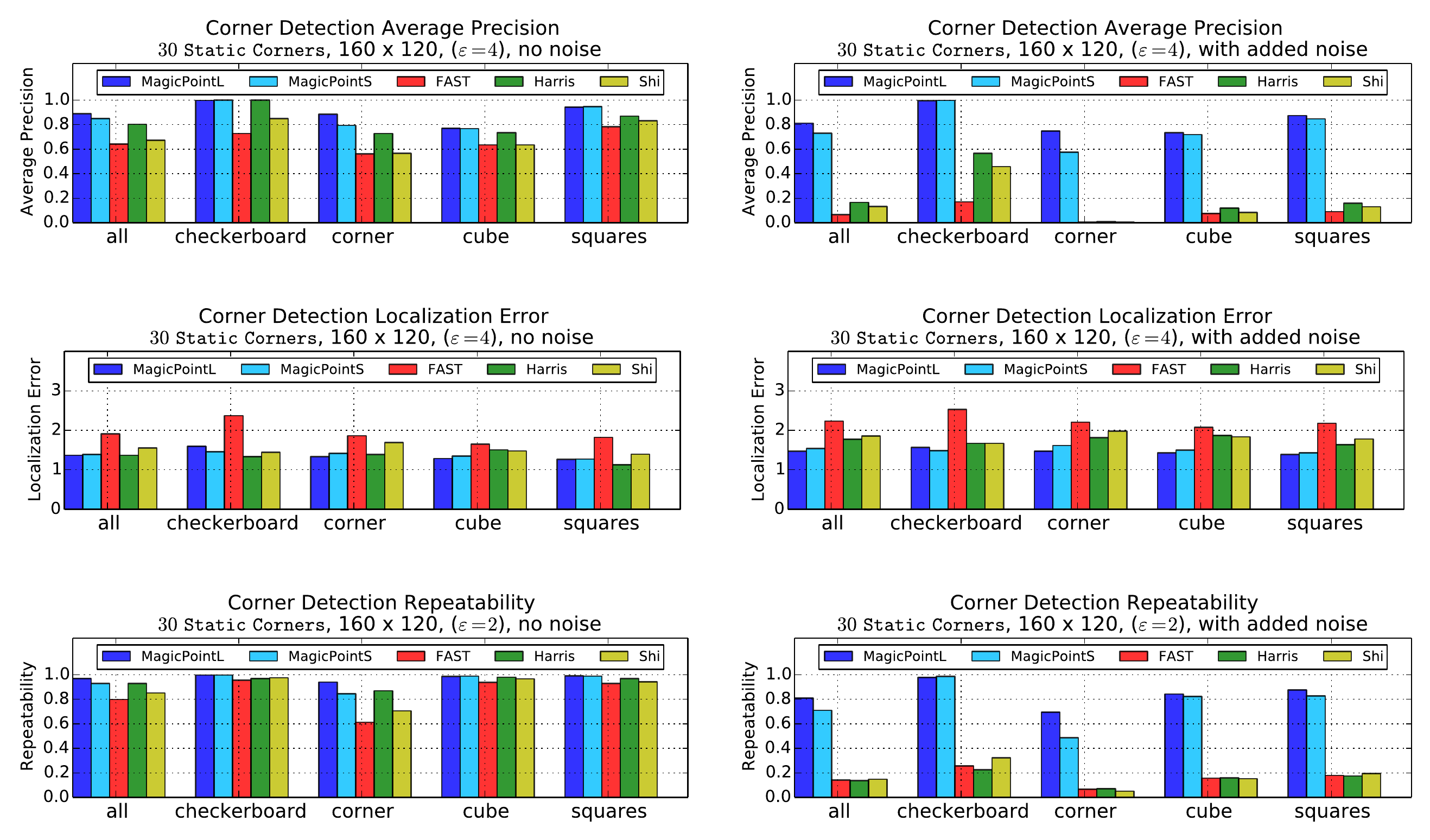}
\caption{{\bf 30 Static Corners Results Plots.} We report three metrics for the detectors on real video sequences with varying lighting conditions. \label{fig:mpeval_sc}}
\end{figure}

\textbf{Mean Average Precision, Mean Localization Error and Repeatability.} For each of the 30 sequences in the dataset, we compute Average Precision, Localization Error and Repeatability (metrics are described in detail in Section~\ref{sec:measures}) with and without noise. The results are broken down by corner category in Figure \ref{fig:mpeval_sc}. We are able to measure Repeatability in this dataset because we now have a sequence of images viewing the same scene in each frame. We see a similar story as we did in the {\tt Synthetic Shapes} evaluation. The MagicPoint detectors detect more corners more confidently and with better localization, especially in the presence of noise. The corners are also more repeatable across frames, showing their robustness to lighting variation.

\begin{figure}
\centering
\includegraphics[width=\textwidth]{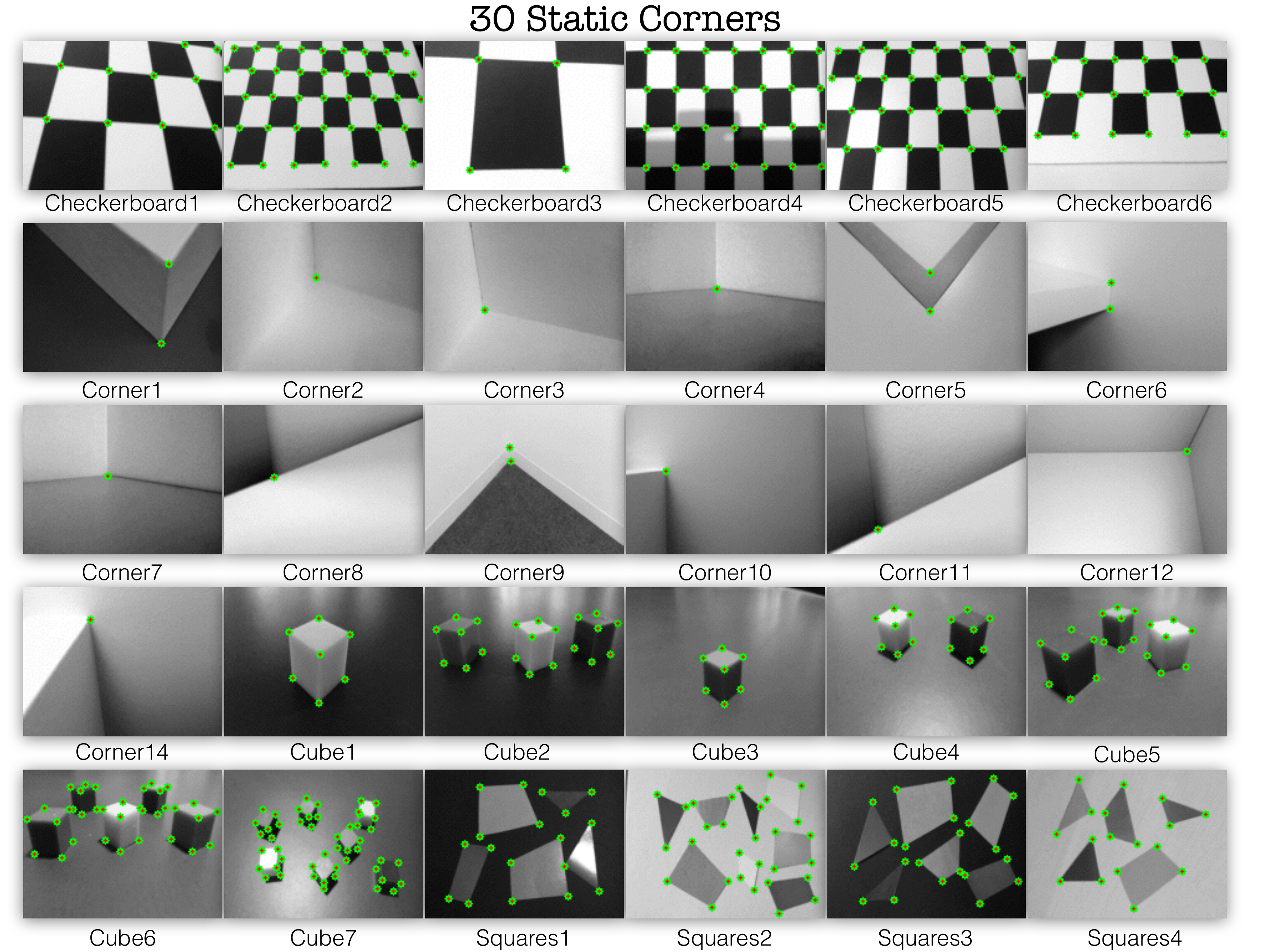}
\caption{{\bf 30 Static Corners Dataset.} We show example frames for each sequence in the {\tt 30 Static Corners} dataset alongside the ground truth corner locations. The sequences come from four different categories: checkerboards, isolated corners, cubes, and squares.  \label{fig:sc}}
\end{figure}

\textbf{Effect of Image Size.} The experiments reported above were conducted at a 160x120 image resolution, which is the same resolution that we used to train the MagicPoint detector. This resolution is smaller than the resolution used in most modern SLAM systems and probably smaller than what the classical detectors were designed for. Since the MagicPoint detector is a fully-convolutional model, we can run it at different input resolutions. We repeated the above {\tt 30 Static Corners} experiments at the 320x240 resolution, and report the results in Table \ref{table:mptable_sc}. The MagicPointL detector outperforms the other models in every setting except for no noise localization error, in which the Harris detector scores the best. However, in the presence of noise, MagicPointL and MagicPointS score the best. It is expected for manually designed corner detectors to work best in ideal conditions, however, engineered feature detectors prove to be too brittle in the presence of noise.

\textbf{Compute Analysis.} For an input image size 160x120, the average forward pass times on a single CPU for MagicPointS and MagicPointL are 5.3ms and 19.4ms respectively. For an input image size of 320x240, the average forward pass times on a single CPU for MagicPointS and MagicPointL are 38.1ms and 150.9ms respectively. The times were computed with BatchNorm layers folded into the convolutional layers.

\begin{table}
\begin{center}
 \begin{tabular}{| l | l | l | c c c c c|}
 \hline
 Metric & Noise& Resolution & MagicPointL & MagicPointS & FAST & Harris & Shi\\
 \hline\hline
 mAP & no & 160x120 & \textbf{0.888} & 0.850 & 0.642 & 0.803 & 0.674 \\
 \hline
 mAP & yes & 160x120 & \textbf{0.811} & 0.730 & 0.066 & 0.166 & 0.132 \\
 \hline
 MLE & no & 160x120 & \textbf{1.365} & 1.391 & 1.908 & 1.369 & 1.551 \\
 \hline
 MLE & yes & 160x120 & \textbf{1.470} & 1.537 & 2.236 & 1.775 & 1.858 \\
 \hline
 R & no & 160x120 & \textbf{0.970} & 0.929 & 0.800 & 0.929 & 0.852 \\
 \hline
 R & yes & 160x120 & \textbf{0.811} & 0.711 & 0.141 & 0.136 & 0.148 \\
 \hline
 mAP & no & 320x240 & \textbf{0.892} & 0.816 & 0.405 & 0.678 & 0.686 \\
 \hline
 mAP & yes & 320x240 & \textbf{0.846} & 0.687 & 0.018 & 0.072 & 0.077 \\
 \hline
 MLE & no & 320x240 & 1.455 & 1.450 & 1.914 & \textbf{1.438} & 1.592 \\
 \hline
 MLE & yes & 320x240 & \textbf{1.533} & 1.605 & 2.200 & 1.764 & 1.848 \\
 \hline
 R & no & 320x240 & \textbf{0.959} & 0.920 & 0.812 & 0.896 & 0.827 \\
 \hline
 R & yes & 320x240 & \textbf{0.765} & 0.675 & 0.099 & 0.081 & 0.104 \\
 \hline
\end{tabular}
\end{center}
\caption{\textbf{Static Corners Results Table.} Reports the mean Average Precision (mAP, higher is better), Mean Localization Error (MLE, lower is better) and Repeatability (R, higher is better) across the 30 real data sequences.}
\label{table:mptable_sc}
\end{table}

\section{MagicWarp Evaluation}

MagicWarp is designed to operate on top of a fast, geometrically stable point detector running in a tracking scenario. In an ideal world, the underlying point detector would be so fast and powerful that a simple nearest neighbor approach would be sufficient to establish correspondence across frames. We believe that MagicPoint is step the right direction in this regard, but it is not yet perfect, and some logic is still required to clean up the mistakes made by the point detector and occlusions between detections.

On this premise, we devised an evaluation for MagicWarp in which points are randomly placed in an image and undergo four simple transformations. ``Translation'' is a simple right translation. ``Rotation'' is an in-plane rotation. ``Scale'' is a zoom-in operation. ``Random H'' is a more complex motion that samples a random homography in which the average displacement of points at the corners of the image is 30 pixels in a 160x120 image. Transformations are applied to various densities of point images and various amounts of extra random points added to to the point set $x_j$. The Nearest Neighbor baseline uses the 3x3 identity matrix $I$ for $H$ and MagicWarp uses the 3x3 matrix output from the network for $H$.

To measure the performance of MagicWarp, we compute a Match Correctness percentage. More specifically, given a set of points $x_i$ in one image where ${i \in \{1,\dots,N_1\}}$ we define a ground truth transformation ${\hat{H}}$ and the predicted transformation $H$. We also define the set of points in the second image as $x_j$ where ${i \in \{1,\dots,N_2\}}$. Match Correctness determines if a transformed point $x'_i$ has a correct nearest neighbor.
\begin{equation}
  \verb+MatchCorr+({\bf x}_i) = (\argmin_{j \in \{1,\dots,N_2\}} ||H{{\bf x}'_i} - {\bf x}_j||) ==  \hat{H}{{\bf x'}_i}
  \label{eqn:repeatability-match}
\end{equation}
Match Repeatability counts the percentage of correct matches made.
\begin{equation}
  \verb+MatchRep+ = 100*\frac{1}{N_1} (\sum_{i} \verb+MatchCorr+({\bf x}_i))
\end{equation}

Table \ref{table:phtable} aims to answer the question: how extreme of a transformation can the correspondence algorithm handle? To answer this question, we linearly interpolate between the identity transformation and each of the four transformations described above and measure the point at which the Match Repeatability drops less than 90\%. We choose 90\% because we believe that a robust geometric decomposition using the correspondences should be able to deal with 10\% of incorrect matches. Unsurprisingly, the MagicWarp approach outperforms the Nearest Neighbor matching approach in all scenarios.

\begin{table}
\setlength\tabcolsep{2.5pt} 
\begin{tabular}{| c | c || C | C | C | C || C | C | C | C |}
\hline
\multicolumn{2}{|c||}{} & \multicolumn{4}{|c||}{Nearest Neighbor} & \multicolumn{4}{|c|}{MagicWarp}\\
\hline
Point Density & Noise  & \mathrm{Trans} &\mathrm{Rot}  &\mathrm{Scale}  &\mathrm{RandH}   &\mathrm{Trans}  &\mathrm{Rot}    &\mathrm{Scale}  &\mathrm{RandH} \\
\hline

\multirow{3}{*}{Low [5,25]}  &0\%  &8.41\mathrm{px} &9.42^{\circ} &1.20\times &13.89\mathrm{px} &\bf{24.00px} &\bf{21.45}^{\circ} &\bf{1.32\times} &\bf{32.83px} \\
  					         &20\% &9.05\mathrm{px} &8.87^{\circ} &1.24\times &11.76\mathrm{px} &\bf{24.06px} &\bf{21.25}^{\circ} &\bf{1.31\times} &\bf{29.78px} \\
                             &40\% &7.15\mathrm{px} &7.70^{\circ} &1.20\times &11.59\mathrm{px} &\bf{22.64px} &\bf{19.65}^{\circ} &\bf{1.20\times} &\bf{28.84px} \\
\hline
\multirow{3}{*}{Medium [25,50]}  &0\%  &5.19\mathrm{px} &5.93^{\circ} &1.11\times &8.03\mathrm{px}  &\bf{20.20px} &\bf{20.01}^{\circ} &\bf{1.23\times} &\bf{26.52px} \\
                                 &20\% &4.82\mathrm{px} &5.41^{\circ} &1.11\times &7.68\mathrm{px}  &\bf{18.29px} &\bf{18.07}^{\circ} &\bf{1.21\times} &\bf{24.84px} \\
                                 &40\% &4.66\mathrm{px} &4.50^{\circ} &1.10\times &6.56\mathrm{px}  &\bf{17.08px} &\bf{18.03}^{\circ} &\bf{1.19\times} &\bf{24.47px} \\
\hline
\multirow{3}{*}{High [100,200]} &0\%  &3.49\mathrm{px} &3.49^{\circ} &1.07\times &4.87\mathrm{px} &\bf{15.10px} &\bf{15.38}^{\circ} &\bf{1.17\times}  &\bf{17.51px} \\
                                &20\% &3.39\mathrm{px} &3.34^{\circ} &1.06\times &4.72\mathrm{px} &\bf{12.97px} &\bf{13.97}^{\circ} &\bf{1.13\times}	&\bf{15.35px} \\
                                &40\% &3.27\mathrm{px} &3.13^{\circ} &1.06\times &4.26\mathrm{px} &\bf{10.92px} &\bf{11.17}^{\circ} &\bf{1.10\times}	&\bf{12.13px} \\
\hline

\end{tabular}
\caption{\textbf{Matching Algorithm 90\% Breakdown Point Experiment}. This table compares matching ability of MagicWarp to a Nearest Neighbor matching approach. Each table entry is the magnitude of transformation that results in fewer than 90\% Match Repeatability across a pair of input points. Higher is better and the MagicWarp approach performs best in all scenarios. Values are averaged across 50 runs.}
\label{table:phtable}
\end{table}

\begin{figure*}
\centering
\includegraphics[width=\textwidth]{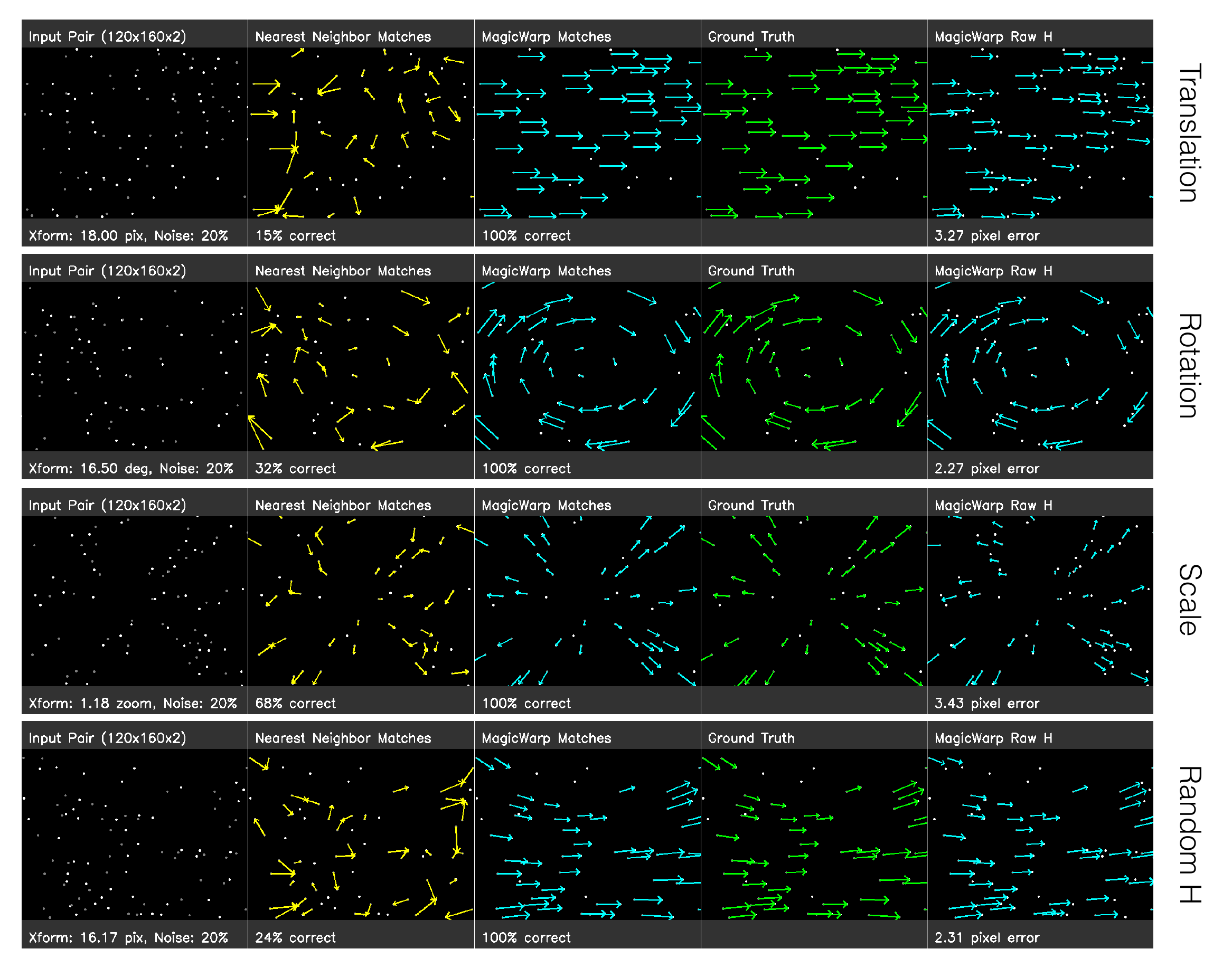}
\caption{\textbf{MagicWarp In Action.} Examples of MagicWarp for each of the four transformation types summarized in Table~\ref{table:phtable}. The left-most column shows the point image input pair overlayed onto a single image. The right-most column shows the MagicWarp's raw predicted homography applied to the gray point set. The middle column shows the MagicWarp result which applies nearest neighbor to this raw predicted homography, which snaps the arrows to the correct points.}
\label{fig:phexamples}
\vspace{-.2in}
\end{figure*}

\begin{figure*}
\centering
\includegraphics[width=\textwidth]{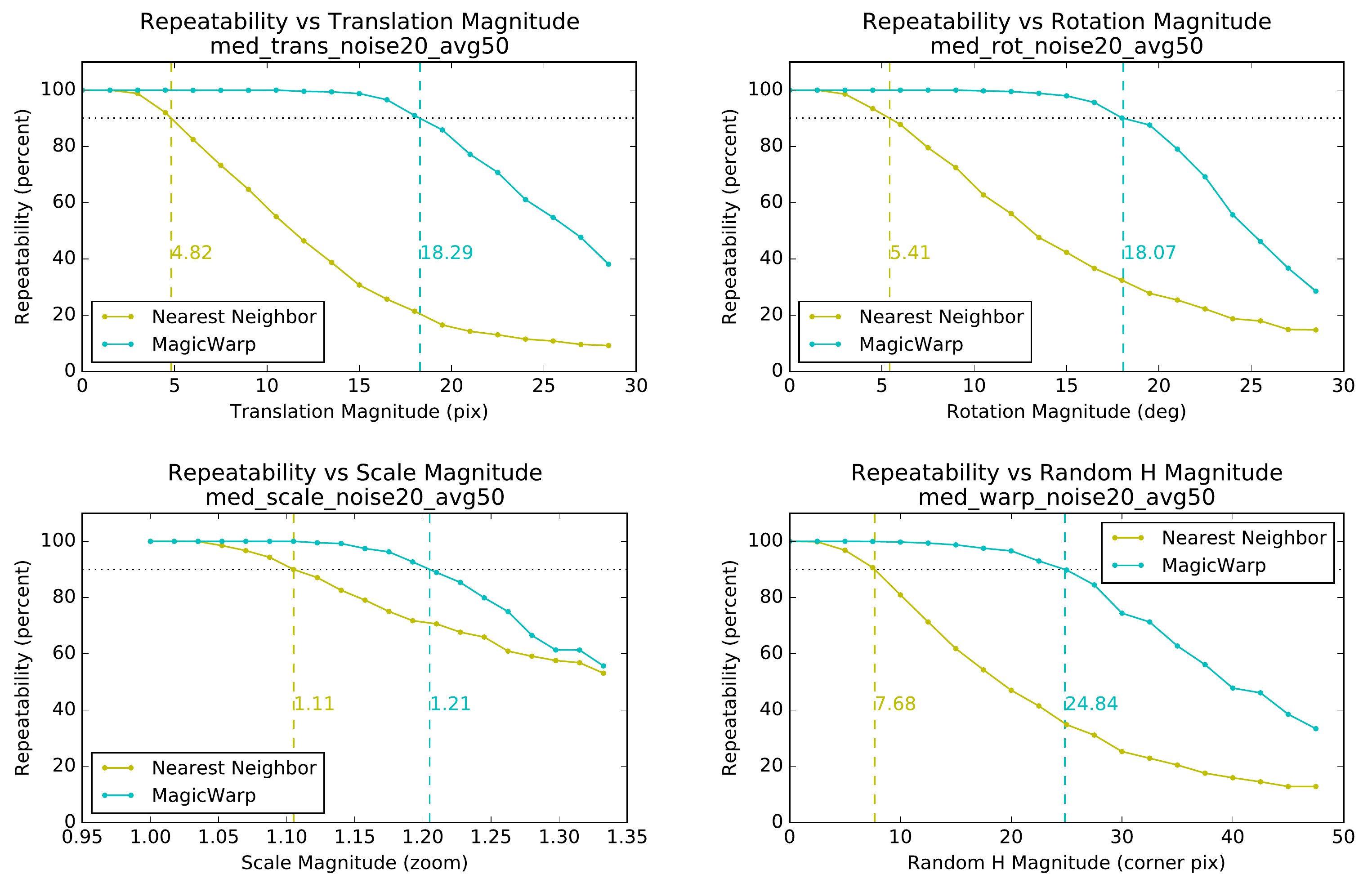}
\caption{\textbf{MagicWarp Average Match Repeatability.} Match Repeatability is compared versus transformation magnitude for four types of transformations. The point image pairs have medium density and 20\% noise added. The vertical dashed lines show the breakdown points at 90\%, which are summarized for different configurations in Table \ref{table:phtable}}
\label{fig:phplots}
\vspace{-.2in}
\end{figure*}

\begin{figure*}
\centering
\includegraphics[width=\textwidth]{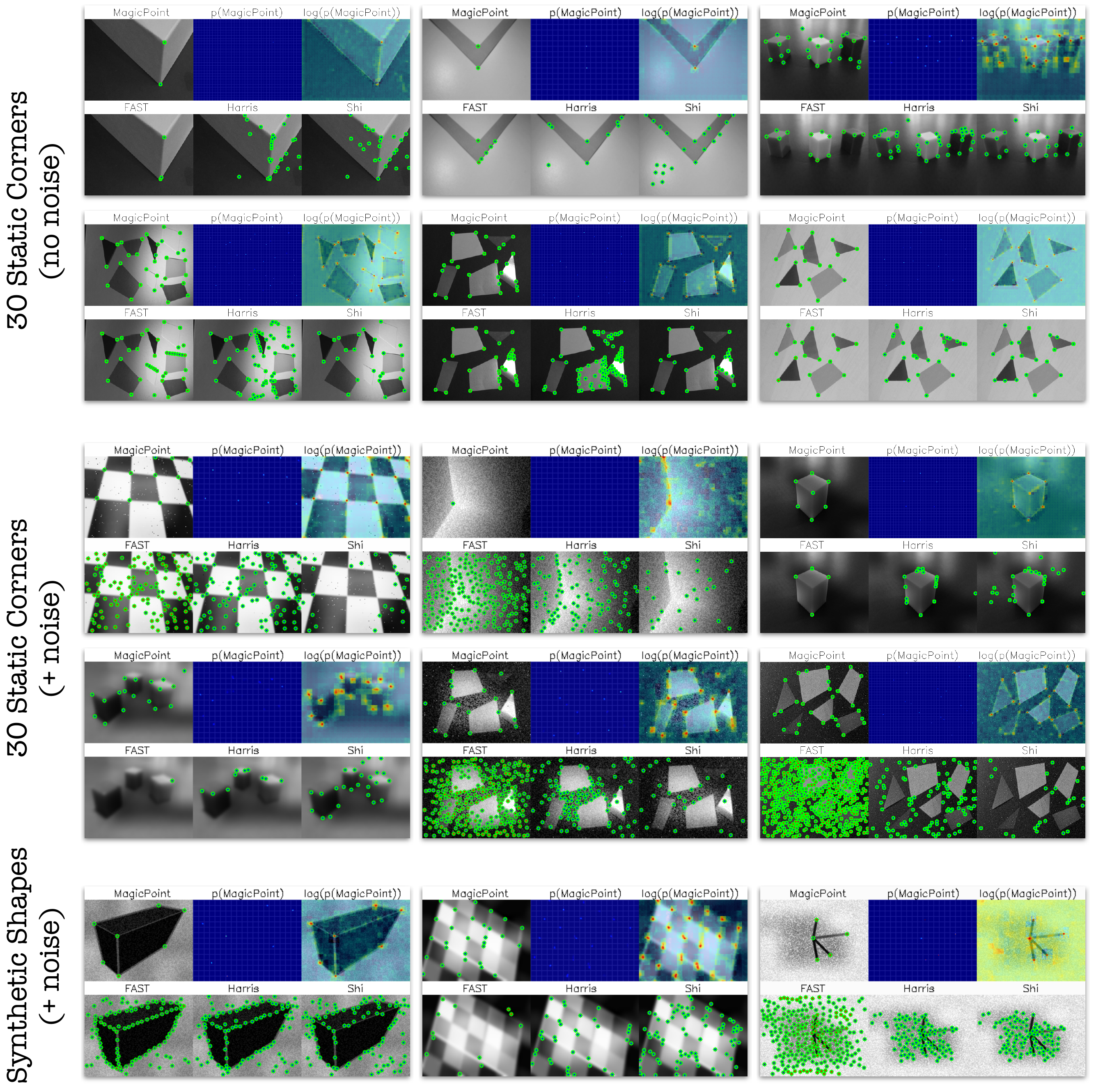}
\caption{\textbf{MagicPoint in Action.} This figures show 15 example results for MagicPointS vs traditional corner detection baselines. For each figure, we display the MagicPointS output, the output probability heatmap, the overlayed log(probability) heatmap (to enhance low probabilities), as well as FAST, Harris, and Shi. The top examples are from {\tt 30 Static Corners} with no noise. The middle examples are from {\tt 30 Static Corners} with noise. The bottom examples are from {\tt Synthetic Shapes} with noise. Note that our method is able to cope with large amounts of noise and produces meaningful heatmaps that can be thresholded in an application-specific manner.}
\label{fig:mp_action}
\vspace{-.2in}
\end{figure*}

MagicWarp is very efficient. For an input size of 20x15x130 (corresponding to an image size of 160x120), the average forward pass time on a single CPU is 2.3 ms. For an input size of 40x30x130 (corresponding to an image size of 320x240), the average forward pass time on a single CPU is 6.1 ms. The times were computed with BatchNorm layers folded into the convolutional layers.

\vspace{-.1in}
\section{Discussion}
\vspace{-.1in}
\label{sec:discussion}

In conclusion, our contributions are as follows. We formulated two SLAM subtasks as machine learning problems, developed two simple data generators which can be implemented in a few hundred lines of code, designed two simple convolutional neural networks capable of running in real-time, and evaluated them on both synthetic and real data.

Our paper was motivated by two burning questions: 1.) \emph{What would it take to build an ImageNet-scale dataset for SLAM?} and 2.) \emph{What would it take to build DeepSLAM?} In this paper, we have shown that our answers to both questions are intimately related. It would be wasteful to build a massive dataset first, only to learn a year later that the best algorithm does not even use the labels you worked so hard to procure. We started with a mental framework for Deep Visual SLAM that must solve two separate subtasks, which can be combined into a point tracking system. By moving away from full frame prediction and focusing solely on geometric consistency, our work has hopefully shown that the day of ImageNet-sized SLAM datasets might not need to come, after all. We believe that the day of massive-scale deployment of Deep-Learning powered SLAM systems is not far.



\clearpage


{\small
\bibliography{example}}  

\end{document}